\documentclass[conference]{IEEEtran}
\IEEEoverridecommandlockouts
\usepackage{cite}
\usepackage{amsmath,amssymb,amsfonts}
\usepackage{algorithm}
\usepackage{algorithmic}
\usepackage{graphicx}
\usepackage[inline]{enumitem}
\usepackage{textcomp}
\usepackage{mytodonotes}
\usepackage{xcolor}
\usepackage{amsthm}
\usepackage{xurl}
\usepackage[hidelinks]{hyperref}
\usepackage{cleveref}
\usepackage{myacronyms}
\usepackage{subcaption}
\theoremstyle{plain}
\newtheorem{assumption}{Assumption}
\crefname{assumption}{assumption}{assumptions}
\Crefname{assumption}{Assumption}{Assumptions}
\theoremstyle{definition}
\newtheorem{definition}{Definition}
\crefname{definition}{definition}{definitions}
\Crefname{definition}{Definition}{Definitions}
\def\BibTeX{{\rm B\kern-.05em{\sc i\kern-.025em b}\kern-.08em
    T\kern-.1667em\lower.7ex\hbox{E}\kern-.125emX}}
\begin{document}

\newlist{questions}{enumerate}{2}
\setlist[questions,1]{label=(RQ\arabic*),ref=RQ\arabic*,leftmargin=*}
\setlist[questions,2]{label=(\alph*),ref=\thequestionsi(\alph*)}

\newcommand{\approach}{\textsc{C²FL}}
\title{\approach{}: Clustered Continual Federated Learning under Spatial and Temporal Drift}

\author{

\IEEEauthorblockN{Davide Domini}
\IEEEauthorblockA{
\textit{University of Bologna}\\
Cesena, Italy  \\
davide.domini@unibo.it}
\and
\IEEEauthorblockN{Gianluca Aguzzi}
\IEEEauthorblockA{
\textit{University of Bologna}\\
Cesena, Italy  \\
gianluca.aguzzi@unibo.it}
\and
\IEEEauthorblockN{Lorenzo Pellegrini}
\IEEEauthorblockA{
\textit{University of Bologna}\\
Cesena, Italy  \\
l.pellegrini@unibo.it}
\and
\IEEEauthorblockN{Mirko Viroli}
\IEEEauthorblockA{
\textit{University of Bologna}\\
Cesena, Italy  \\
mirko.viroli@unibo.it}
\and
\IEEEauthorblockN{Lukas Esterle}
\IEEEauthorblockA{
\textit{Aarhus University}\\
Aarhus, Denmark  \\
lukas.esterle@ece.au.dk}

}
\maketitle

\begin{abstract}
\ac{CAS} increasingly rely on machine learning to let each node learn from locally sensed data, 
aligning its behavior with the surrounding environment. 
Scaling this intelligence, however, 
 raises fundamental challenges: 
 sensed data is often privacy-sensitive, preventing centralized collection; 
 nodes are mobile, traversing regions where nearby nodes perceive similar phenomena while distant ones observe radically different conditions,  creating natural spatial clusters; 
 and these distributions evolve over time due to mobility, 
 introducing temporal drift that makes local models progressively stale.
These dynamics arise across domains -- 
 vehicular sensing, drone-based monitoring, smartphone crowdsensing -- 
 yet the interplay of privacy, spatial heterogeneity, 
 and temporal drift severely undermines conventional learning strategies.
Therefore, we propose \approach{}, 
 a fully distributed \ac{FL} approach where nodes self-organize into learning groups through spatial clustering, 
 reflecting the geographic structure of the environment.
To counteract temporal drift, 
 each node combines experience replay with a dwell-time-aware adaptive averaging step,
 progressively incorporating the regional consensus as it remains longer within the same area,
 while preserving previously acquired knowledge under evolving distributions. 
We evaluate our approach on synthetic experiments that systematically reproduce spatial and temporal shifts, 
 showing that standard federated strategies degrade significantly under these conditions 
 and that our method restores robust collective adaptation.
\end{abstract}


\section{Introduction}\label{sec:intro}
Recent advancements in decentralized computing pave the way for embedding intelligence directly at the network edge,
 enabling entities of \acf{CAS} to autonomously tune their operations using on-device machine learning. 
In domains characterized by persistent mobility -- 
 such as connected vehicular networks, 
 drone swarms, 
 and participatory crowdsensing -- 
 these autonomous nodes must perpetually refine their understanding of the environment.
However, 
 accomplishing this adaptation while respecting strict privacy boundaries and limited communication bandwidth introduces substantial hurdles. 
 Consequently, standard distributed learning methodologies frequently prove inadequate for sustaining this pervasive edge intelligence.

\begin{figure}
    \centering
    \includegraphics[width=0.55\columnwidth]{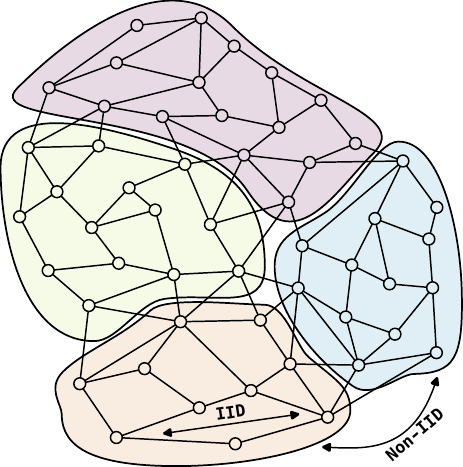}
    \caption{
    Proximity-based heterogeneous data distribution. 
    Circles represent devices, lines represent communication links between devices, 
    and background colors indicate different data distributions. 
    Data are homogeneous within subregions and non-IID across subregions.
    }
    \label{fig:subregions}
\end{figure}

A primary source of difficulty stems from the \emph{spatial} structure of the observed data. 
 Devices that are geographically close typically perceive similar phenomena and therefore collect data with comparable statistical properties, 
 whereas nodes located in distant regions observe radically different conditions~\cite{DBLP:journals/sensors/AzevedoEDMRA22,DBLP:journals/corr/abs-2410-15693}. 
This induces a form of structured heterogeneity in which data is approximately homogeneous within local subregions 
 but strongly non-Independent and Identically Distributed (non-IID) across the wider environment (see~\Cref{fig:subregions}). 
When nodes move across these regions, this spatial heterogeneity acquires a \emph{temporal} dimension: 
 each device is exposed to a sequence of locally coherent but globally distinct data regimes, 
 making its local distribution inherently non-stationary. 
Over time, these mobility-induced shifts can make previously trained models stale 
 and degrade performance on earlier contexts through catastrophic forgetting~\cite{DBLP:journals/nn/ParisiKPKW19}.

At the same time, the data sensed at the network edge is often privacy-sensitive, 
 preventing straightforward centralized collection and training. 
\acf{FL} has therefore emerged as a natural paradigm for this setting, 
 allowing devices to collaboratively train a shared model without exchanging raw data~\cite{DBLP:journals/ijon/LiuLGL24}. 
FL has been successfully deployed in large-scale distributed environments~\cite{DBLP:journals/corr/abs-1811-03604, DBLP:journals/ojcomps/DuWYYJL20,10.1109/TMC.2022.3157603,FANG2024105006}, 
 and has recently been applied to collective self-adaptive systems~\cite{DBLP:journals/iot/DominiFAVE26}. 
To better handle spatial non-IID data, prior work has proposed \acf{CFL} approaches~\cite{DBLP:journals/tit/GhoshCYR22}, 
 which group devices with similar distributions and train a separate model for each cluster. 
While effective in static populations, these methods largely ignore the inherent mobility of devices in CASs 
 and therefore do not directly address the sequential distribution shifts experienced by moving nodes. 
 Addressing this limitation is essential to design federated algorithms that can retain prior knowledge while adapting to novel environments encountered during mobility.

To close this gap, in this work, 
we propose \approach{} (Clustered Continual Federated Learning), a decentralized cluster-based continual federated learning approach
 for mobile spatial environments. \approach{} builds on the FBFL framework~\cite{DBLP:journals/lmcs/DominiAEV26}
 to preserve decentralized spatial clustering, while incorporating \ac{CL} mechanisms
 to cope with the sequential distribution shifts induced by mobility~\cite{DBLP:journals/nn/ParisiKPKW19}. 
While recent studies have begun to explore the intersection of \ac{FL},
 \ac{CL}, and mobility~\cite{DBLP:journals/corr/abs-2411-13740}, to the best of our knowledge no prior
 work has explicitly combined decentralized coordination, \ac{CFL},
 and continual adaptation within a unified framework. 
CL methods are specifically designed to enable models to adapt 
to sequentially changing data distributions while mitigating catastrophic forgetting, 
for instance by leveraging experience replay together with progressive model integration across successive regimes.
In particular, we instantiate this design with replay and an adaptive averaging mechanism,
which gradually blends the regional consensus into the device model as permanence in the current region increases, and
 experimentally evaluate their behavior in a federated setting characterized by spatially structured
 data heterogeneity and device mobility. Our experiments leverage widely used benchmark datasets
 synthetically partitioned to create challenging scenarios for evaluating the proposed approach.

The remainder of this paper is organized as follows. 
\Cref{sec:related} reviews the relevant background and related work.
\Cref{sec:rqs} presents the motivation of this work and formulates the research questions. 
\Cref{sec:formalization} formalizes the considered system model and makes explicit how node mobility induces a continual task stream. 
\Cref{sec:approach} discusses the proposed \approach{} approach.
\Cref{sec:eval} reports the experimental evaluation and discusses the obtained results. 
Finally, 
 \Cref{sec:conclusion} concludes the paper and outlines directions for future research.

\section{Background and Motivation}\label{sec:related}

\subsection{Federated Learning}

\acf{FL}~\cite{DBLP:conf/aistats/McMahanMRHA17,DBLP:journals/corr/McMahanMRA16} 
 has emerged as a distributed machine learning paradigm 
 that enables multiple clients to collaboratively train a shared model
 while keeping data locally on devices, 
 thus preserving privacy. 
In the standard formulation, 
 each client performs local training and periodically shares model updates with a central server, 
 which aggregates them to produce a global model. 
This process is iterated over multiple communication rounds.

A key assumption underlying classical FL is that client data are 
 independent and identically distributed (IID).
However, 
 this assumption rarely holds in real-world deployments. 
In practice, 
 data across clients are typically non-IID due to variations in user behavior, 
 environmental conditions, or sensing contexts. 
This statistical heterogeneity is known to negatively affect convergence and model performance~\cite{DBLP:journals/csur/ChungLLCH26,DBLP:journals/ijon/ZhuXLJ21}, 
 often leading to model drift and unstable optimization dynamics.
To address this issue, 
 a line of research has focused on \acl{CFL}~\cite{liu2025survey}, 
 where the objective is no longer to learn a single global model, 
 but rather to partition clients into groups based on similarity in their data distributions 
 and train a separate model for each cluster. 
This approach enables a form of implicit personalization, 
 improving performance in heterogeneous settings by restricting collaboration 
 to statistically similar clients. 

Moreover, 
 in \ac{CAS} and large-scale IoT environments, 
 data heterogeneity often exhibits a 
 spatial structure~\cite{DBLP:journals/jors/DominiIALV26}. 
Devices that are geographically close tend to observe similar phenomena, 
 leading to locally homogeneous data distributions, 
 while distant devices may experience significantly different conditions. 
This results in \emph{proximity-based non-IID data} (\Cref{fig:subregions}), 
 where heterogeneity emerges across spatial regions rather than randomly across clients. 

Another fundamental challenge in applying FL to CAS lies in the reliance on centralized coordination. 
Traditional server-based architectures introduce scalability bottlenecks, 
 communication overhead, 
 and single points of failure, 
 which are particularly problematic in highly distributed 
 and dynamic environments such as IoT systems. 
To overcome these limitations, 
 recent works have proposed \emph{self-organizing} and decentralized FL 
 approaches~\cite{DBLP:journals/iot/DominiFAVE26,DBLP:journals/lmcs/DominiAEV26}, 
 where clients autonomously form collaboration groups and perform learning without a central coordinator. 
These methods improve scalability and resilience by enabling adaptive, 
 multi-model learning through local interactions. 
However, 
 existing approaches typically assume relatively static network conditions, 
 limiting their applicability in highly dynamic scenarios characterized by mobility 
 and evolving topologies. 

\subsection{Continual Learning}
Continual Learning (\emph{CL}), 
 sometimes referred to as \emph{lifelong learning}, 
 studies learning systems that are exposed to a sequence of tasks or data distributions over time, 
 rather than to a single stationary training set~\cite{DBLP:journals/nn/ParisiKPKW19}. 
The goal is to enable a model to acquire new knowledge while preserving previously learned capabilities. 
 This setting naturally arises in long-lived adaptive systems, 
 where the operating conditions of a device may evolve due to changes in the environment, 
 user behavior, task requirements, or mobility. 
In such scenarios, 
 retraining a model from scratch whenever new data becomes available is often impractical, 
 either because data from past experiences is no longer accessible, 
 or because compute, time, 
 or other resource constraints make a full retraining infeasible. 
The main challenge addressed by \ac{CL} is \emph{catastrophic forgetting}, 
 namely the tendency of neural models to quickly lose performance on previously learned tasks when optimized on new ones~\cite{french1999catastrophic}.

Depending on how the data stream evolves, 
 \ac{CL} problems are commonly distinguished into task-incremental, domain-incremental, and class-incremental settings~\cite{vandeven2022three}. 
 In \emph{task-incremental} learning, 
 the model observes a sequence of distinct tasks, 
 often with task identifiers available at training and inference time, 
 which serve as a way to partition a global problem into different parts. 
In \emph{domain/data-incremental} learning, the input distribution changes while the prediction space remains fixed. 
In \emph{class-incremental} learning, 
 new classes are progressively introduced, 
 and the model must discriminate among all classes observed so far. 
Beyond this basic taxonomy, 
 other scenarios have been proposed to better capture the properties of realistic streams. 
 For instance, \emph{class-incremental learning with repetitions} allows previously observed classes to reappear later in the stream, 
 modeling the fact that real environments often revisit past concepts rather than presenting each class only once~\cite{hemati2023class}.

To mitigate forgetting, 
 different methods have been proposed, which are usually grouped into three categories:
 \begin{enumerate*}[label=(\roman*)]
     \item \emph{regularization} approaches constrain the update of model parameters to preserve knowledge relevant to previous tasks; 
     \item \emph{replay} approaches store past examples and interleave them with current data~\cite{lopezpaz2017gradient,rebuffi2017icarl}; 
     \item \emph{architectural} methods allocate task-specific model components, for instance, through dynamic expansion, masking, or task-specific modules.
 \end{enumerate*} 
 
Among these strategies, replay is particularly relevant for distributed edge settings,
since it can be performed locally without sharing raw data.
In mobile spatial environments, replay can be complemented by lightweight parameter-integration rules
that progressively mix the regional consensus into the device model as the node stabilizes in a region.
For mobile nodes in spatial environments,
 this combination provides a practical way to adapt to newly visited regions while retaining knowledge acquired from regions visited earlier in time. 
 On the other hand, architectural methods often rely on explicit task signals or well-defined task boundaries~\cite{rusu2016progressive,yoon2018lifelong,mallya2018packnet,mallya2018piggyback}, 
 which can make them less suitable for continuous adaptation in decentralized settings where task boundaries may be blurred.

\subsection{Federated Continual Learning}

\ac{FCL}, 
 which combines \ac{FL} and \ac{CL}, 
 has emerged as a promising approach to handle non-stationary data in distributed systems,
 enabling models to adapt over time while preserving privacy. 
Existing approaches can be broadly categorized based on how 
 they mitigate catastrophic forgetting in federated settings~\cite{CRIADO2022263}, 
 including:
\begin{enumerate*}[label=(\roman*)]
    \item regularization-based methods;
    \item replay or memory-based techniques;
    \item model-integration strategies; and
    \item architectural decomposition schemes.
\end{enumerate*}
Surveys such as \cite{DBLP:journals/ijon/HamediRH25,DBLP:journals/corr/abs-2411-13740} 
 highlight that these methods address different variants of continual learning 
 (e.g., task-, domain-, or class-incremental) under federated constraints, 
 and are increasingly motivated by edge and mobile applications characterized 
 by heterogeneous and evolving data distributions.

Several works explicitly consider dynamic environments where data distributions 
 change over time due to mobility or concept drift~\cite{DBLP:journals/mta/CasadoLCIRB22,DBLP:journals/corr/abs-2111-07457}. 
For instance, 
 extensions of FedAvg incorporate drift detection and adaptive updates 
 to cope with non-stationary data, 
 while other approaches introduce attention mechanisms or personalized model components 
 to improve adaptation in rapidly changing settings. 
More advanced frameworks, 
 such as Cross-\ac{FCL}~\cite{DBLP:journals/tmc/ZhangGSLY24}, 
 tackle scenarios where devices move across multiple edge systems, 
 proposing parameter decomposition and cross-edge knowledge transfer 
 to balance adaptation and retention. 
Similarly, 
 lifelong federated learning~\cite{DBLP:conf/euspn/YuQW22} 
 has been explored in mobile robotics, 
 where agents continuously acquire new knowledge while collaborating 
 through federated updates.
In parallel, 
 mobility-aware \ac{FL}~\cite{DBLP:journals/corr/abs-2108-09103} 
 research has shown that client movement both exacerbates and, 
 in some cases, alleviates statistical heterogeneity. 
Mobility can degrade convergence due to intermittent participation and shifting data, 
 but can also act as a mechanism for implicit data mixing 
 and knowledge propagation across the network~\cite{DBLP:journals/tvt/ChenYSZGN25}. 
This dual role has motivated approaches that explicitly model mobility 
 in the learning process, 
 for example through hierarchical aggregation, mobility-aware sampling, 
 or decentralized coordination strategies. 
However, 
 these works typically focus on improving convergence and communication efficiency, 
 rather than addressing continual adaptation at the model level.

Despite these advances, 
 existing \ac{FCL} approaches largely assume either centralized coordination, 
 static or weakly dynamic client groupings, 
 or ignore the spatial structure of data altogether. 
While some works combine \ac{FL} and \ac{CL} to handle non-stationarity, 
 they do not explicitly consider clustered or proximity-based learning, 
 and conversely, \ac{CFL} approaches generally assume stable client memberships 
 and do not incorporate continual learning mechanisms. 

\subsection{Motivation and Research Questions}\label{sec:rqs}

Existing literature still lacks a unified treatment of three aspects that 
 jointly characterize mobile \ac{CAS} scenarios: 
 decentralized clustered coordination, 
 spatially structured data heterogeneity, 
 and continual adaptation 
 to mobility-induced distribution shifts. 
This limitation is particularly relevant when nodes move across regions, 
 because each device encounters a sequence of locally coherent but distinct 
 data regimes, 
 which can degrade performance over time and induce catastrophic 
 forgetting. 
Mobility-aware \ac{FL} methods mainly target convergence and communication 
 efficiency, 
 whereas current \ac{FCL} approaches do not explicitly account for 
 decentralized clustered coordination in spatially structured environments.
This motivates our work on a unified framework that combines decentralized 
 coordination, 
 \ac{CFL}, and \ac{CL} to study whether such an integration can 
 better support adaptation under mobility-induced drift.

We therefore investigate the following research questions:
\begin{questions}
    \item Does node mobility in decentralized clustered federated learning induce sequential distribution shifts that lead to performance degradation and catastrophic forgetting?\label{itm:rq1}
    \item Can a decentralized approach that balances local adaptation with global model integration,
     enriched with continual learning mechanisms, improve knowledge retention and robustness under mobility-induced drift?\label{itm:rq2}
\end{questions}

\section{Problem Formulation}\label{sec:formalization}

\begin{table}[t]
\centering
\caption{Summary of the main notation used in the paper.}
\label{tab:notation}
\footnotesize
\renewcommand{\arraystretch}{1.4}
\begin{tabular}{lp{0.62\columnwidth}}
\hline
\textbf{Symbol} & \textbf{Meaning} \\
\hline
$V$ & Set of devices \\
$T$ & Number of learning rounds \\
$S$ & Spatial domain \\
$S_k$ & $k$-th spatial subregion \\
$K$ & Number of subregions \\
$\mathbf{p}^{(t)}_d$ & Position of device $d$ at round $t$ \\
$r^{(t)}_d$ & Region of device $d$ at round $t$ \\
$\mathcal{R}^{(t)}_k$ & Devices in region/cluster $k$ \\
$G^{(t)}=(V,E^{(t)})$ & Communication graph \\
$\mathcal{N}^{(t)}_d$ & Neighbors of device $d$ \\
$\mathcal{D}^{(t)}_d$ & Local dataset of device $d$ \\
$n^{(t)}_d$ & Number of local samples \\
$\mathbb{P}^{(t)}_d$ & Local data distribution \\
$\epsilon_1,\epsilon_2$ & Intra-/inter-region thresholds \\
$\theta^{(t)}_d$ & Device model parameters \\
$\mathcal{L}^{(t)}_d$ & Local empirical objective \\
$\mathcal{J}^{(t)}_k$ & Region-wise objective \\
$\mathcal{T}^{(t)}_d$ & Task of device $d$ at round $t$ \\
$\mathcal{H}^{(t)}_d$ & Task history of device $d$ \\
$\eta^{(t)}_d$ & Elected cluster leader \\
$g^{(t)}_d$ & Distance-to-leader field \\
$R$ & Cluster radius \\
$\widetilde{\theta}^{(t+1)}_d$ & Locally trained model \\
$\theta^{(t+1)}_k$ & Regional consensus model \\
$\mathcal{M}^{(t)}_d$ & Replay memory \\
$s^{(t)}_d$ & Dwell-time counter \\
$H$ & Averaging horizon \\
$\Gamma$ & Maximum mixing factor \\
$\alpha^{(t)}_d$ & Adaptive mixing factor \\
$\mathrm{Acc}^{(t)}_d(k)$ & Per-region accuracy \\
$\mathrm{CAcc}^{(t)}$ & Cumulative accuracy \\
\hline
\end{tabular}
\end{table}

This section formalizes the system model and the learning problem addressed in this work;
 the main symbols used throughout the paper are summarized in~\Cref{tab:notation}.
Specifically, we characterize how node mobility inherently induces a continual task stream for a set of mobile devices $\mathcal{V}$, 
which collaboratively learn over a discrete time horizon $T$, i.e., for rounds $t \in \{1,\dots,T\}$,
while navigating a spatial environment characterized by structured data heterogeneity.
To this end, we first detail the spatial environment and the mobility model (\Cref{sec:formalization-mobility}), 
followed by a formalization of the region-induced data heterogeneity (\Cref{sec:formalization-data}). 
Finally, we define the learning objectives and the resulting mobility-induced continual task stream (\Cref{sec:formalization-objective,sec:continual-learning-stream}).

\paragraph*{Notation}
Throughout the remainder of the paper, 
we use superscripts to denote the time index (e.g., $\mathbf{p}_d^{(t)}$ for the position of device $d$ at round $t$) 
and subscripts for the device index (e.g., $\mathcal{D}_d^{(t)}$ for the local dataset of device $d$ at round $t$). 
Boldface fonts (e.g., $\mathbf{p}$, $\mathbf{x}$) represent vectors and matrices, 
while regular fonts are used for scalars (e.g., $t$, $d$).

\subsection{Spatial Environment and Mobility}\label{sec:formalization-mobility}
Let $\mathcal{S} \subseteq \mathbb{R}^2$ be a spatial domain partitioned into $K$ disjoint subregions
 $\{\mathcal{S}_k\}_{k=1}^K$ such that $\bigcup_{k=1}^K \mathcal{S}_k = \mathcal{S}$.
Each device $d \in \mathcal{V}$ occupies a position $\mathbf{p}_d^{(t)} \in \mathcal{S}$ at round $t$.
We denote by $r_d^{(t)} \in \{1, \dots, K\}$ the index of the region containing device $d$ at time $t$, namely
\[
    r_d^{(t)} = k \quad \Longleftrightarrow \quad \mathbf{p}_d^{(t)} \in \mathcal{S}_k.
\]
Hence, the trajectory $\{\mathbf{p}_d^{(t)}\}_{t=1}^T$ induces for each device a region-membership sequence $\{r_d^{(t)}\}_{t=1}^T$.
For each region $k$, we write
\[
    \mathcal{R}_k^{(t)} = \{d \in \mathcal{V} : r_d^{(t)} = k\}
\]
for the set of devices currently located in $\mathcal{S}_k$.

The communication pattern is modeled as a time-varying graph
 $\mathcal{G}^{(t)} = (\mathcal{V}, \mathcal{E}^{(t)})$,
 where $(d,d') \in \mathcal{E}^{(t)}$ if and only if devices $d$ and $d'$ can directly exchange information at round $t$.
We denote by $\mathcal{N}_d^{(t)} = \{d' \in \mathcal{V} : (d,d') \in \mathcal{E}^{(t)}\}$ the communication neighborhood of node $d$.
Node mobility therefore affects the system along two coupled dimensions:
 it changes the local environment from which a device collects data,
 and it changes the set of peers with which the same device can collaborate.

\subsection{Region-Induced Data Heterogeneity}\label{sec:formalization-data}
Let $\mathcal{X}$ and $\mathcal{Y}$ denote the input and output spaces, respectively.
Specifically, $\mathcal{X}$ represents the space of observations locally perceived by the devices (e.g., environmental sensor readings), while $\mathcal{Y}$ represents the corresponding target variables to be predicted.
We denote outputs by $\mathbf{y} \in \mathcal{Y}$; scalar targets are included as the one-dimensional special case.
At round $t$, each device $d$ is associated with a local dataset
 $\mathcal{D}_d^{(t)} = \{(\mathbf{x}_{d,i}^{(t)}, \mathbf{y}_{d,i}^{(t)})\}_{i=1}^{n_d^{(t)}}$
sampled from a local distribution $\mathbb{P}_d^{(t)}(\mathbf{x},\mathbf{y})$ over $\mathcal{X} \times \mathcal{Y}$, where $n_d^{(t)}$ denotes the number of samples observed by device $d$ at round $t$.
When the time index is implicit, 
we simply write $\mathcal{D}_d$ and $\mathbb{P}_d$.
Let $D(\cdot,\cdot)$ denote a statistical distance between probability distributions.
Since data is generated by the surrounding physical context, 
$\mathbb{P}_d^{(t)}$ depends on the region currently occupied by the device.

\begin{assumption}[Spatial-Data Correlation]\label{asm:spatial-correlation}
Devices located in the same geographic region exhibit similar data distributions.
Specifically, for any region $h \in \{1, \dots, K\}$ and any two devices $d, d' \in \mathcal{R}_h^{(t)}$,
\[
    D\bigl(\mathbb{P}_d^{(t)}, \mathbb{P}_{d'}^{(t)}\bigr) \le \epsilon_1,
\]
for an intra-region similarity threshold $\epsilon_1 > 0$.
\end{assumption}

\begin{assumption}[Inter-Region Heterogeneity]\label{asm:heterogeneity}
Data distributions across different subregions are non-IID.
For any two distinct regions $i \neq j$, and any devices $d \in \mathcal{R}_i^{(t)}$ and $d' \in \mathcal{R}_j^{(t)}$,
\[
    D\bigl(\mathbb{P}_d^{(t)}, \mathbb{P}_{d'}^{(t)}\bigr) \ge \epsilon_2,
\]
where $\epsilon_2 > \epsilon_1$ establishes a strict separation gap between inter-region heterogeneity and intra-region similarity.
\end{assumption}

Given these assumptions,
 when a device moves across regions,
 its local data distribution changes abruptly,
 turning spatial heterogeneity into temporal drift along its trajectory.

\subsection{Region-Wise Objective}\label{sec:formalization-objective}
Let $f_{\theta} : \mathcal{X} \to \mathcal{Y}$ be a prediction model parameterized by $\theta \in \mathbb{R}^p$, 
 and let $\ell\bigl(f_{\theta}(\mathbf{x}), \mathbf{y}\bigr)$ be the per-sample loss (e.g., cross-entropy for classification, mean squared error for regression).
We define the local empirical objective of device $d$ at round $t$ as
\[
    \mathcal{L}_d^{(t)}(\theta)
    := \frac{1}{n_d^{(t)}} \sum_{i=1}^{n_d^{(t)}}
    \ell\bigl(f_\theta(\mathbf{x}_{d,i}^{(t)}), \mathbf{y}_{d,i}^{(t)}\bigr).
\]
For each region $k \in \{1,\dots,K\}$ and round $t$, let
\[
    N_k^{(t)} := \sum_{d \in \mathcal{R}_k^{(t)}} n_d^{(t)}
\]
denote the total number of samples currently available in region $k$.
We then define the region-wise empirical objective
\[
    \begin{aligned}
        \mathcal{J}_k^{(t)}(\theta)
        &:= \frac{1}{N_k^{(t)}}
        \sum_{d \in \mathcal{R}_k^{(t)}} \sum_{i=1}^{n_d^{(t)}}
        \ell\bigl(f_\theta(\mathbf{x}_{d,i}^{(t)}), \mathbf{y}_{d,i}^{(t)}\bigr) \\
        &= \frac{1}{N_k^{(t)}} \sum_{d \in \mathcal{R}_k^{(t)}} n_d^{(t)} \mathcal{L}_d^{(t)}(\theta).
    \end{aligned}
\]
That is, 
$\mathcal{J}_k^{(t)}(\theta)$ is simply the average loss over all samples currently observed by the devices located in region $k$, which equivalently corresponds to the weighted average of the local empirical objectives.
An ideal region-specific model therefore satisfies
\[
    \theta_k^{(t)\star} \in \arg\min_{\theta \in \mathbb{R}^p} \mathcal{J}_k^{(t)}(\theta).
\]
In the decentralized setting considered here, 
 clustered federated learning seeks distributed approximations of these region-wise optima using only local communication from the graph $\mathcal{G}^{(t)}$.

\subsection{Mobility-Induced Continual Learning Stream}\label{sec:continual-learning-stream}
Mobility implies that each device is exposed to a sequence of different local data regimes over time.
To formalize this, we define
\[
    \mathcal{T}_d^{(t)} := \bigl(\mathbb{P}_d^{(t)}, \ell\bigr)
\]
as the local task experienced by device $d$ at round $t$, 
 namely the learning problem induced by the distribution currently observed at that device.

\begin{definition}[Mobility-induced task stream]\label{def:task-stream}
For each device $d \in \mathcal{V}$, the trajectory $\{\mathbf{p}_d^{(t)}\}_{t=1}^T$ induces the task sequence $\{\mathcal{T}_d^{(t)}\}_{t=1}^T$.
We denote by $\mathcal{H}_d^{(t)} = \{\mathcal{T}_d^{(1)}, \dots, \mathcal{T}_d^{(t)}\}$ the task history experienced by device $d$ up to round $t$, and refer to the full sequence $\mathcal{H}_d^{(T)}$ as its continual task stream.
\end{definition}

According to \Cref{def:task-stream}, 
 node mobility transforms a spatial learning problem into a sequential one:
 at round $t$, 
 the same device starts from its current model $\theta_d^{(t)}$ and performs a local adaptation step while traversing different region-specific data regimes.
This local continual-learning update can be abstracted as:
\[
    \theta_d^{(t+1)}
    \approx
    \operatorname{Update}\!\left(
        \theta_d^{(t)},
        \mathcal{L}_d^{(t)}(\theta) + \lambda \,\Omega\bigl(\theta;\mathcal{H}_d^{(t-1)}\bigr)
    \right),
\]
where $\operatorname{Update}(\theta_0,F)$ denotes a few local optimization steps on objective $F$ starting from initialization $\theta_0$. 
Intuitively, $\mathcal{L}_d^{(t)}$ pushes the model to fit the data currently observed by device $d$, 
while $\Omega$ acts as a memory-preserving term that discourages the device from forgetting what it learned in previously visited regions.

\subsection{Overall Decentralized Learning Goal}
Clustered federated learning and continual learning address two complementary dimensions of the same problem, 
namely the spatial coordination of devices currently exposed to similar conditions (collective aspect) 
and the temporal robustness of the models they carry while moving across regions (individual aspect). 
Accordingly, 
the overall problem can be stated as follows: 
design a decentralized learning process that, at each round $t$, 
simultaneously approximates the collection of region-wise objectives $\{\mathcal{J}_k^{(t)}\}_{k=1}^K$ under the communication constraints induced by $\mathcal{G}^{(t)}$ 
and preserves model competence over the device-level task stream $\mathcal{H}_d^{(T)}$. 
Equivalently, 
the desired process should track region-wise optima while limiting catastrophic forgetting in mobile devices.

\section{Clustered Continual Federated Learning: \approach{}}\label{sec:approach}

\approach{} is an interplay of three main components:
\begin{enumerate*}[label=(\roman*)]
\item a decentralized clustering mechanism that groups devices based on the similarity of their local data distributions,
\item a federated learning algorithm that trains cluster-specific models through collaborative optimization,
and \item a continual learning module that enables each device to adapt to sequential distribution shifts while mitigating catastrophic forgetting.
\end{enumerate*}
For sake of clarity, 
 we present these components as distinct modules,
 with the same temporal ordering as the one described above.
However, 
 in practice, these processes may have different timescales and may be interleaved in various ways---more details on this aspect are discussed in \Cref{sec:results}.
The complete protocol for one communication round is formalized in \Cref{alg:c2fl}.

\paragraph{Decentralized Clustering}
In \approach{}, the spatial clustering of mobile devices is achieved through a self-stabilizing leader election process implemented using the \emph{aggregate computing} paradigm~\cite{DBLP:journals/computer/BealPV15}. 
Specifically, 
 we leverage the \emph{S-building block} (Sparse-choice) pattern~\cite{DBLP:conf/saso/MoBD18}, 
 which partitions the network into contiguous regions, each centered around a dynamically elected leader. 

Formally, 
 each node $d \in \mathcal{V}$ maintains a local state $\sigma_d^{(t)} = \langle \eta_d^{(t)}, g_d^{(t)} \rangle$, 
  where $\eta_d^{(t)} \in \mathcal{V}$ identifies its currently elected cluster leader and $g_d^{(t)} \in \mathbb{R}_{\ge 0}$ tracks the shortest communication distance to it. 
 The process is driven by a persistent priority $v_k \in \mathbb{R}$ assigned to each node $k$ (e.g., based on its ID or a random value), which determines the leadership hierarchy and ensures convergence. 
 At every round $t$, nodes exchange their states with their neighbors $\mathcal{N}_d^{(t)}$, 
 and each node $d$ updates its state through the following emergent logic:

\begin{enumerate}
    \item \emph{Information Integration}: 
    Node $d$ constructs a set of perceived leadership claims $\mathcal{C}_d^{(t)}$ by relaying advertisements from its neighbors and considering its own potential leadership:
    \begin{equation}
        \small
        \mathcal{C}_d^{(t)} = \{ (\eta_{d'}^{(t-1)}, g_{d'}^{(t-1)} + \|\mathbf{p}_d^{(t)} - \mathbf{p}_{d'}^{(t)}\|) \mid d' \in \mathcal{N}_d^{(t)} \} \\ \cup \{ (d, 0) \}
    \end{equation}
    where $\|\mathbf{p}_d^{(t)} - \mathbf{p}_{d'}^{(t)}\|$ is the Euclidean distance to neighbor $d'$. Each claim represents a candidate leader and the current estimated path length to reach it.
    \item \emph{Candidate Pruning}: 
    To ensure that clusters reflect local spatial correlation, 
    node $d$ filters $\mathcal{C}_d^{(t)}$ to keep only those candidates reachable within a maximum \emph{radius of influence} $R$, 
    forming the candidate set $\mathcal{K}_d^{(t)} = \{ (k, g) \in \mathcal{C}_d^{(t)} \mid g \le R \}$.
    \item \emph{Priority-Based Election}: The node elects the ``best'' candidate from $\mathcal{K}_d^{(t)}$ by maximizing the priority $v_k$.
    The state is then updated as:
    \begin{equation}
        \langle \eta_d^{(t)}, g_d^{(t)} \rangle = \text{argmax}_{\langle k, g \rangle \in \mathcal{K}_d^{(t)}} \{ (v_k, k) \}
    \end{equation}
\end{enumerate}

Assuming the radius $R$ is chosen appropriately with respect to the spatial distribution of the nodes, 
this mechanism yields a Voronoi-like partition where each node $d$ belongs to a $\mathcal{R}_k^{(t)}$ (we will use $\mathcal{R}_k^{(t)}$ to denote both the region and the corresponding cluster of devices)
centered around its leader $\eta_d^{(t)}$, 
which lives inside $k$'s radius of influence and is separated from other clusters by a boundary of width at least $R$, 
capturing the spatial correlation structure of the data. 
We refer to $\eta_{d \to k}^{(t)}$ as the cluster leader of node $d$ in region $k$ at round $t$.
The logic is \emph{self-stabilizing}: 
 if a node $d$ moves beyond its current leader's radius or a leader fails, 
 the corresponding claims are pruned or naturally expire, 
 triggering a re-election that seamlessly realigns the clusters to the new topology. 
This ensures that learning groups remain consistent with the underlying data heterogeneity~(\Cref{asm:spatial-correlation,asm:heterogeneity}) 
 without central coordination.

\paragraph{Decentralized Clustered-Based Federated Learning}
The federated learning process operates within the established clusters, 
  utilizing the FBFL paradigm~\cite{DBLP:journals/lmcs/DominiAEV26} (rooted to the SCR pattern~\cite{DBLP:journals/fgcs/PianiniCVN21}) to build regional consensus. 
  This process is orchestrated through two complementary communication patterns from the aggregate computing literature: 
  \emph{collect-cast} (or C-block)~\cite{DBLP:conf/atal/AudritoBDV19} for aggregation and \emph{gradient-cast} (or G-block) 
  for dissemination~\cite{DBLP:conf/saso/AudritoCDV17}.

The learning cycle for each cluster $\mathcal{R}_k^{(t)}$ consists of the following steps:
\begin{enumerate}
    \item \emph{Local Training}: 
    Each device $d \in \mathcal{R}_k^{(t)}$ performs a fixed number of local gradient descent steps via an optimization algorithm (e.g., Adam) on its effective local training set (potentially augmented with replayed samples from previous regions, as described in the next section),
    starting from the local model $\theta_d^{(t)}$. 
    This yields an updated local model $\widetilde{\theta}_d^{(t+1)}$.
    \item \emph{Consensus Aggregation (Collect-cast)}: 
    The locally trained models are routed towards the cluster leader $\eta_{d \to k}^{(t)}$ (simplified as $\eta_d^{(t)}$ for sake of simplicity)
    by following the gradient field $g_d^{(t)}$ in reverse. 
    Formally, 
    information flows from nodes with larger $g_d^{(t)}$ to those with smaller values, 
    effectively performing a spatial reduction over the cluster topology. 
    The leader $\eta_d^{(t)}$ collects these models and computes the new regional consensus using a weighted average:
    \begin{equation}
        \theta_k^{(t+1)} = \frac{1}{N_k^{(t)}} \sum_{d \in \mathcal{R}_k^{(t)}} n_d^{(t)} \, \widetilde{\theta}_d^{(t+1)}
    \end{equation}
    where $N_k^{(t)}$ is the total sample count in the cluster.
    \item \emph{Model Dissemination (Gradient-cast)}: 
    Once the consensus $\theta_k^{(t+1)}$ is computed, 
    the leader broadcasts it back to all members of $\mathcal{R}_k^{(t)}$. 
    This dissemination follows the distance field $g_d^{(t)}$, 
    where the leader acts as the source and the model parameters propagate ``downhill'' to reach every node within the radius $R$.
\end{enumerate}
The resulting dissemination and aggregation processes are inherently decentralized, 
 allowing the system to continuously adapt to the dynamic cluster topology induced by node mobility. 
However, 
 this fully distributed coordination introduces intrinsic communication latency, 
 as locally trained models and regional consensus may require multiple interaction steps to propagate across the cluster via multi-hop communication. 
To address this, 
 each device $d$ performs a local \emph{knowledge merging} step, 
 reconciling the asynchronously received regional consensus with its current local state before continuing the learning process.

\paragraph{Continual Learning via Adaptive Averaging and Replay}


\approach{} relies on two complementary mechanisms, namely \emph{replay} and \emph{adaptive averaging}, to mitigate forgetting while allowing the model to adapt to new regions.
First, 
the generic memory-preserving component $\Omega(\theta;\mathcal{H}_d^{(t-1)})$ is realized implicitly through \emph{Replay}: 
rather than adding an explicit regularization term in parameter space, the node augments the current local data with samples collected in previously visited regions.
Denoting
\[
\begin{aligned}
    \widehat{\mathcal{L}}_d^{(t)}(\theta)
    := &
    \frac{1}{\lvert\mathcal{M}_d^{(t)}\rvert + n_d^{(t)}} \\
    & \cdot \Biggl(
        \sum_{(\mathbf{x},\mathbf{y}) \in \mathcal{D}_d^{(t)}} \ell\bigl(f_\theta(\mathbf{x}), \mathbf{y}\bigr) \\
        & \quad + \sum_{(\mathbf{x},\mathbf{y}) \in \mathcal{M}_d^{(t)}} \ell\bigl(f_\theta(\mathbf{x}), \mathbf{y}\bigr)
    \Biggr),
\end{aligned}
\]
the replay-augmented local objective, then the \emph{Local Training} phase can be written as
\[
    \widetilde{\theta}_d^{(t+1)}
    \approx
    \operatorname{Update}\!\bigl(\theta_d^{(t)}, \widehat{\mathcal{L}}_d^{(t)}(\theta)\bigr),
\]
which is the concrete realization adopted in \approach{} of the abstract form
$\mathcal{L}_d^{(t)}(\theta) + \lambda\,\Omega(\theta;\mathcal{H}_d^{(t-1)})$, 
with the historical contribution encoded by replayed samples.
Concretely, the node optimizes on $\mathcal{M}_d^{(t)} \cup \mathcal{D}_d^{(t)}$, 
 where $\mathcal{D}_d^{(t)}$ is the latest local dataset perceived in the current region and $\mathcal{M}_d^{(t)}$ stores data collected before the current region. 
 The replay memory is updated only when the device moves to a different region:
\[
    \mathcal{M}_d^{(t)}
    =
    \begin{cases}
        \mathcal{M}_d^{(t-1)}, & \text{if } \eta_d^{(t)} = \eta_d^{(t-1)}, \\
        \mathcal{M}_d^{(t-1)} \cup \mathcal{D}_d^{(t-1)}, & \text{otherwise.} \\
        \mathcal{M}_d^{(0)} = \emptyset
    \end{cases}
\]
Thus, the current-region contribution is always given by the latest $\mathcal{D}_d^{(t)}$, 
 whereas the historical replay memory grows only at region changes by appending $\mathcal{D}_d^{(t-1)}$.
In this work, we assume an unbounded replay memory in order to isolate the effect 
of mobility-induced forgetting. An analysis on the impact of bounded-size memory replay policies, 
such as the ones based on class-balanced reservoir sampling \cite{pmlr-v119-chrysakis20a}, is left to future work.
The goal of the replay mechanism is to reduce forgetting of knowledge acquired in
previous regions. 
Second, after the \emph{Model Dissemination} step, 
 the received regional model $\theta_k^{(t+1)}$ is merged into the device state through an adaptive averaging rule.
Let $s_d^{(t)}$ denote the number of consecutive rounds that device $d$ has spent in its current region, namely
\[
    s_d^{(t)} =
    \begin{cases}
        s_d^{(t-1)} + 1, & \text{if } \eta_d^{(t)} = \eta_d^{(t-1)}, \\
        1, & \text{otherwise.}
    \end{cases}
\]
Given an adaptation horizon $H > 0$ and a maximum collective mixing factor of $\Gamma \in [0,1]$
, we define the collective mixing factor as
\[
    \alpha_d^{(t)} = \min\!\left\{\Gamma, \frac{s_d^{(t)}}{H}\right\}.
\]
Denoting by $\widetilde{\theta}_d^{(t+1)}$ the model obtained after local training with replay, then the final device model is updated as
\[
    \theta_d^{(t+1)} = (1 - \alpha_d^{(t)})\,\widetilde{\theta}_d^{(t+1)} + \alpha_d^{(t)}\,\theta_k^{(t+1)}.
\]
This second stage complements the local continual update above: replay preserves knowledge from the device history, while adaptive averaging aligns the resulting model with the current cluster-level consensus.
This rule makes the collective contribution small immediately after a region change and progressively stronger while the node remains in the same area, 
 thereby integrating local-area knowledge without abrupt overwriting.
For static devices, no region transition occurs, 
 so $\mathcal{M}_d^{(t)} = \emptyset$ and local training simply relies on $\mathcal{D}_d^{(t)}$, 
 i.e., the latest local dataset perceived in the current area. 
In this case, the local model progressively aligns with the regional consensus while still benefiting from local updates.
Replay memory $\mathcal{M}_d^{(t)}$ is never shared with other nodes; 
 accordingly, moving devices influence federated aggregation only through the locally trained model contributed during \emph{Consensus Aggregation}. 
 This model already reflects knowledge accumulated in previously visited regions $r_d^{(1\ldots t-1)}$, 
 while the adaptive averaging rule progressively aligns it with the current regional consensus as permanence in the area increases.

\begin{algorithm}[t]
\caption{\approach{}: 
One communication round for device~$d$. 
Here, $\mathcal{D}_d^{(t)}$ is the current local dataset, 
$\eta_d^{(t)}$ is the current cluster leader, 
$g_d^{(t)}$ the distance-to-leader field, 
$s_d^{(t)}$ the permanence counter in the current region, 
$\mathcal{M}_d^{(t)}$ the historical replay memory, 
$\widetilde{\theta}_d^{(t+1)}$ the locally trained model, 
$\theta_k^{(t+1)}$ the regional consensus, 
$\Gamma$ the maximum collective mixing factor,
and $H$ the adaptive-averaging horizon.}
\label{alg:c2fl}
\begin{algorithmic}[1]
\REQUIRE $\mathcal{D}_d^{(t)}$, $\mathcal{D}_d^{(t-1)}$, $\{\sigma_{d'}^{(t-1)}\}_{d' \in \mathcal{N}_d^{(t)}}$, $\theta_d^{(t)}$, $\eta_d^{(t-1)}$, $s_d^{(t-1)}$, $\mathcal{M}_d^{(t-1)}$
\ENSURE $\theta_d^{(t+1)}$, $\mathcal{M}_d^{(t)}$
\STATE \textbf{Phase~1: Decentralized Clustering Update}
\STATE Build $\mathcal{C}_d^{(t)}$ from neighbor claims and self-claim $(d,\,0)$
\STATE $\mathcal{K}_d^{(t)} \leftarrow \{(k,g)\in\mathcal{C}_d^{(t)}: g \le R\}$
\STATE $(\eta_d^{(t)}, g_d^{(t)}) \leftarrow \arg\max_{(k,g)\in\mathcal{K}_d^{(t)}} (v_k,\, k)$
\IF{$\eta_d^{(t)} \neq \eta_d^{(t-1)}$}
\STATE $s_d^{(t)} \leftarrow 1$; $\mathcal{M}_d^{(t)} \leftarrow \mathcal{M}_d^{(t-1)} \cup \mathcal{D}_d^{(t-1)}$
\ELSE
\STATE $s_d^{(t)} \leftarrow s_d^{(t-1)} + 1$; $\mathcal{M}_d^{(t)} \leftarrow \mathcal{M}_d^{(t-1)}$
\ENDIF
\STATE \textbf{Phase~2: Continual Local Training}
\STATE $\widetilde{\theta}_d^{(t+1)} \leftarrow \operatorname{LocalTrain}(\theta_d^{(t)}, \mathcal{M}_d^{(t)} \cup \mathcal{D}_d^{(t)})$
\STATE \textbf{Phase~3: Cluster Aggregation (Collect-cast)}
\STATE Route $\widetilde{\theta}_d^{(t+1)}$ toward $\eta_d^{(t)}$ following decreasing $g_d^{(t)}$
\IF{$d = \eta_d^{(t)}$}
\STATE $\theta_k^{(t+1)} \leftarrow
    \tfrac{1}{N_k^{(t)}}\textstyle\sum_{d'\in\mathcal{R}_k^{(t)}} n_{d'}^{(t)}\,\widetilde{\theta}_{d'}^{(t+1)}$
\ENDIF
\STATE \textbf{Phase~4: Model Dissemination (Gradient-cast)}
\STATE Receive $\theta_k^{(t+1)}$ propagated from $\eta_d^{(t)}$ via increasing $g_d^{(t)}$
\STATE \textbf{Phase~5: Local Knowledge Integration}
\STATE $\alpha_d^{(t)} \leftarrow \min\{\Gamma, s_d^{(t)}/H\}$
\STATE $\theta_d^{(t+1)} \leftarrow (1-\alpha_d^{(t)})\widetilde{\theta}_d^{(t+1)} + \alpha_d^{(t)}\theta_k^{(t+1)}$
\end{algorithmic}
\end{algorithm}

\section{Experimental Evaluation}\label{sec:eval}

\subsection{Experimental Setting}


The experimental evaluation is designed to validate the approach
 introduced in \Cref{sec:approach} in a decentralized learning scenario where
 spatial data heterogeneity and mobility-induced temporal drift occur simultaneously.
To this end,
 we consider a synthetic spatial environment composed of four distinct subregions.
Each subregion is associated with a different data distribution,
 so that devices located within the same area observe statistically similar samples,
 whereas devices belonging to different areas observe non-IID data,
 consistently with the setting illustrated in~\Cref{fig:subregions}
 and with the spatial-data correlation and inter-region heterogeneity assumptions
 formalized in~\Cref{asm:spatial-correlation,asm:heterogeneity}.

We build the learning scenario using the ProFed benchmark~\cite{DBLP:journals/jors/DominiIALV26}
 using the Extended MNIST (EMNIST) dataset and partitioning it into
 multiple synthetic local datasets.
The partitioning follows a proximity-based non-IID scheme:
 each subdataset is assigned to one specific spatial area,
 and devices placed in that area can sense samples only from the corresponding distribution.
As a result, devices in the same region receive homogeneous data,
 while devices in different regions are exposed to heterogeneous data distributions.
This construction allows us to systematically reproduce the spatial heterogeneity
 assumed in~\Cref{sec:formalization-data} and
 to control the distribution shifts experienced by mobile devices.


The simulated system consists of $50$ devices distributed approximately equally across the four subregions.
Each device perceives approximately $200$ local samples per round,
 drawn from the distribution of its current area,
 and trains with a batch size of $32$.
The test set globally covers all four areas,
 with a total of $1600$ samples (i.e., $\sim 400$ per area).
The samples are drawn from the corresponding distribution for each area, 
 so that the test set reflects the same spatial heterogeneity as the training data.
Each device independently executes the protocol described in \Cref{alg:c2fl},
 locally training a multi-layer perceptron (MLP) with two hidden layers of $128$ neurons each,
 ReLU activations, $28$ input units, and $27$ output units.
Local optimization is carried out using the Adam optimizer
 with a learning rate of $0.001$ and cross-entropy loss.
A subset of devices is allowed to move across areas during the learning process,
 following different mobility trajectories.
The fraction of mobile devices is set to $20\%$ of the total population. 
These mobile devices induce temporal drift at the local model level,
 since their data distribution changes whenever they enter a new area.
Static devices, instead,
 remain in their initial area and therefore provide a stable regional learning signal.
The four subregions are indexed $0$ through $3$.
Mobile devices follow a circular trajectory, transitioning to the next area
in the cycle at rounds $30$, $60$, and $90$.
The starting area of each mobile device is chosen independently;
for instance, a device starting in area $0$ visits $0 \to 1 \to 2 \to 3$,
while a device starting in area $2$ visits $2 \to 3 \to 0 \to 1$.
In both cases, each device spends exactly $30$ rounds per region.
The total number of global rounds is set to $120$.

As decentralized federated learning baseline,
 we rely on the self-organizing FBFL~\cite{DBLP:journals/lmcs/DominiAEV26} approach
 implemented through the Phyelds framework~\cite{DBLP:journals/corr/abs-2603-29999}.
This baseline is used with a twofold purpose.
First,
 it provides a reference implementation of decentralized clustered FL
 under proximity-based data heterogeneity.
Second,
 it allows us to explicitly assess whether mobility causes performance degradation
 and catastrophic forgetting when no continual-learning mechanism is introduced.
Starting from this baseline,
 we implement the proposed \approach{} approach as described in~\Cref{sec:approach},
 enriching FBFL with replay-based local training
 and adaptive averaging,
 thereby addressing the overall decentralized learning goal of
 tracking region-wise objectives $\mathcal{J}_k^{(t)}$ while
 preserving competence over the device-level task stream $\mathcal{H}_d^{(T)}$
 formalized in~\Cref{sec:formalization-objective,sec:continual-learning-stream}.


The adaptive averaging mechanism in Phase~5 of \Cref{alg:c2fl}
 is configured with a maximum collective mixing factor of $\Gamma = 0.3$
 and an adaptation horizon of $H = 30$ rounds.
These values govern the adaptive rule
 $\alpha_d^{(t)} = \min\{\Gamma,\, s_d^{(t)}/H\}$:
 with $H = 30$, the mixing factor reaches its cap $\Gamma = 0.3$
 after $9$ consecutive rounds in the same region,
 which aligns with the typical dwell time of mobile devices
 ($30$ rounds per area between consecutive mobility triggers).
The replay memory $\mathcal{M}_d^{(t)}$ is unbounded,
 i.e., every sample encountered in a region is retained
 and made available for local training upon subsequent visits.


We adopt two complementary evaluation metrics: \emph{per-area test accuracy} and \emph{cumulative accuracy}.
The first measures the performance of a device model
 on the test distribution of a single region $k$,
 thereby reflecting how well the model approximates the region-wise accuracy 
 evaluated as the empirical mean over the test set:
$$
    \mathrm{Acc}_d^{(t)}(k) = \frac{1}{\lvert\mathcal{D}_k^{\mathrm{test}}\rvert} \sum_{(\mathbf{x},\mathbf{y}) \in \mathcal{D}_k^{\mathrm{test}}} \mathbf{1}\{f_{\theta_d^{(t)}}(\mathbf{x}) = \mathbf{y}\}
$$
where $\mathcal{D}_k^{\mathrm{test}}$ denotes the held-out test set for region $k$, kept fixed across all rounds and never used during training, and $\mathbf{1}\{\cdot\}$ is the indicator function that takes value $1$ when the model prediction $f_{\theta_d^{(t)}}(\mathbf{x})$ matches the true label $\mathbf{y}$, and $0$ otherwise.
This serves to visualize the impact of mobility-induced task transitions
 on knowledge retention across the device task stream $\mathcal{H}_d^{(T)}$.
While the second aggregates the per-area accuracies into a single scalar measure of overall competence.
We define the cumulative accuracy as the average, over all mobile devices, of the sum of their test accuracies across all regions:
$$
    \mathrm{CAcc}^{(t)} = \frac{1}{\lvert\mathcal{V}_{\mathrm{mob}}\rvert} \sum_{d \in \mathcal{V}_{\mathrm{mob}}} \sum_{k=1}^{K} \mathrm{Acc}_d^{(t)}(k)
$$
where $\mathcal{V}_{\mathrm{mob}}$ denotes the set of mobile devices.
This metric was used as a proxy
 to compare \approach{} against the ablation baselines.


The learning components are implemented in PyTorch~\cite{paszke2017automatic}
All experiments are repeated over $10$ independent random seeds
 in order to reduce the effect of lucky initializations and avoid cherry-picking.
Reported curves and quantitative results correspond
 to averages over these independent runs.
The experimental artifacts,
 including source code, configuration files, and scripts required to reproduce the results,
 are publicly released and permanently archived under a permissive
 license\footnote{\url{https://anonymous.4open.science/r/experiments-2026-ACSOS-CL-for-nodes-movement-in-CAS-53BE/}}.

\subsection{Results}\label{sec:results}

\begin{figure*}
    \centering
    \begin{subfigure}{\columnwidth}
        \centering
        \includegraphics[width=\linewidth]{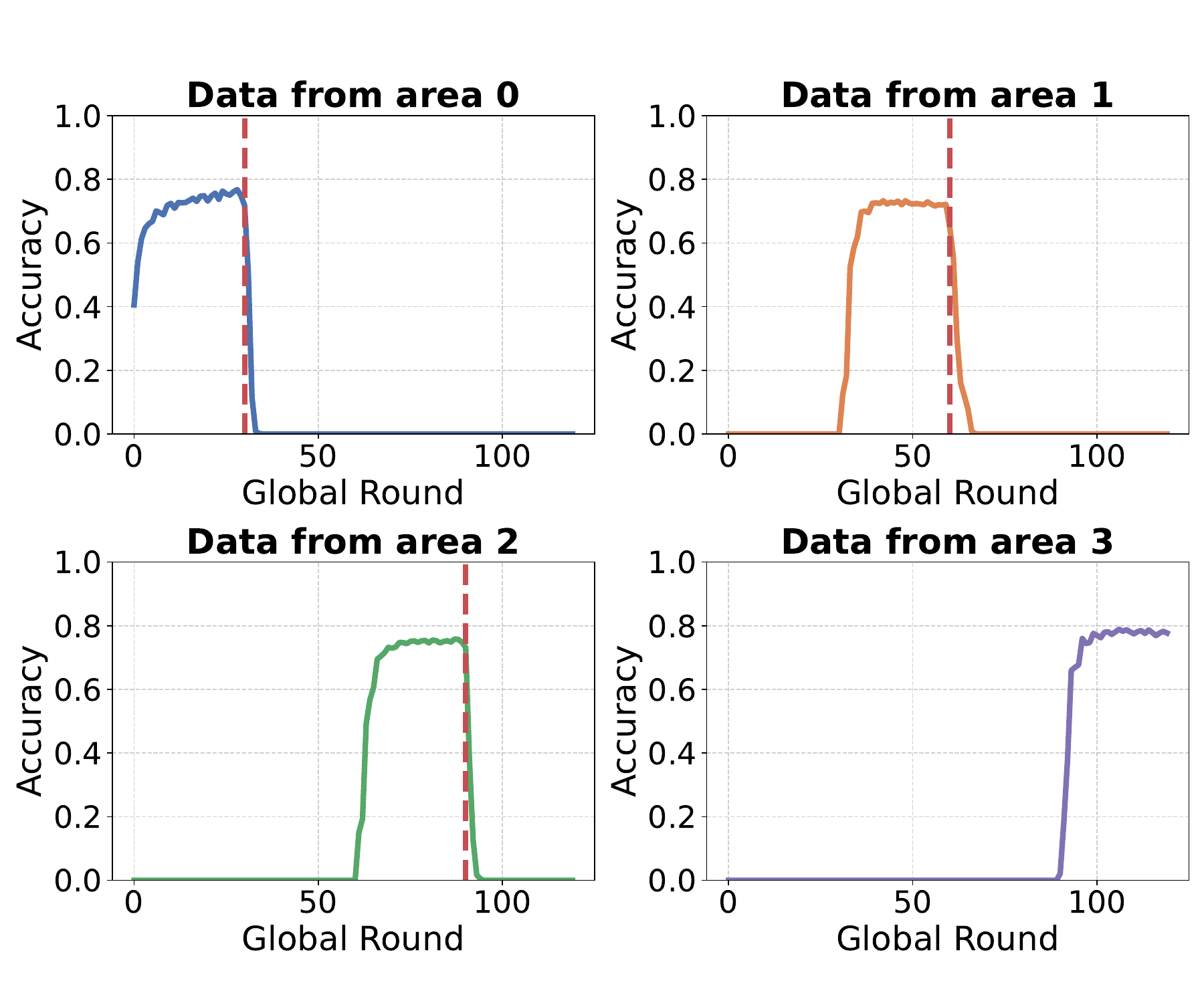}
        \caption{FBFL baseline}
        \label{fig:moving-node-fl}
    \end{subfigure}%
    \begin{subfigure}{\columnwidth}
        \centering
        \includegraphics[width=\linewidth]{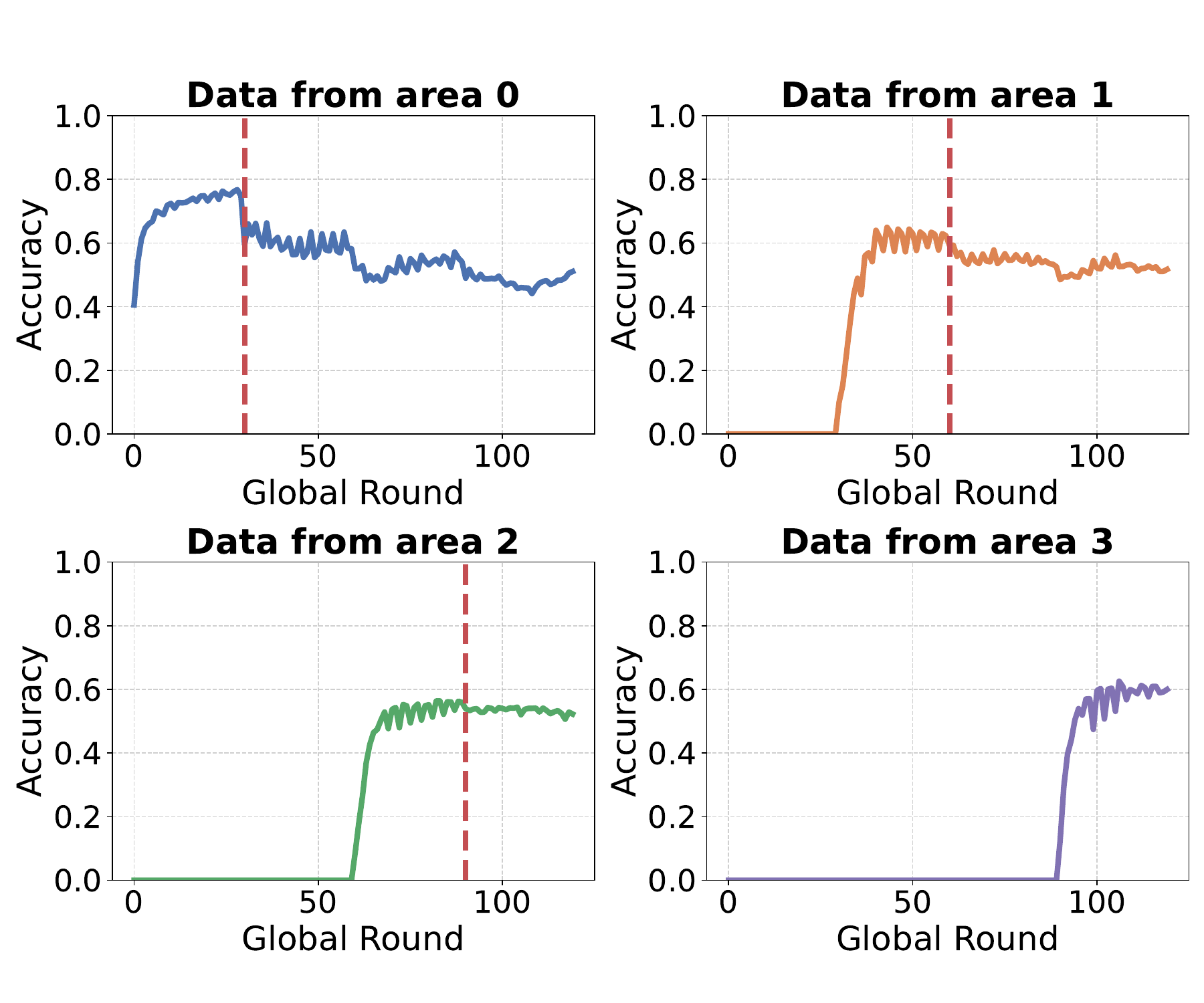}
        \caption{\approach{}}
        \label{fig:moving-node-c2fl}
    \end{subfigure}
    \caption{Per-area accuracy ($\mathrm{Acc}_d^{(t)}(k)$) of a mobile device over the learning process. 
    Each subplot reports the accuracy on the test data associated with one spatial area.
    The vertical dashed red lines mark mobility-induced region transitions.
}
    \label{fig:fl-vs-c2fl}
\end{figure*}

The first experiment evaluates whether mobility alone is sufficient 
 to induce catastrophic forgetting in decentralized clustered federated learning (\ref{itm:rq1}). 
To this end, 
 we run the scenario described in the previous section using the FBFL baseline 
 without any continual-learning mechanism.

\Cref{fig:moving-node-fl} reports the test accuracy of a mobile node 
 over time on the data distributions associated with the four areas. 
Each subplot evaluates the same model on the test data of one specific area, 
 while the vertical dashed red lines indicate the rounds in which the node changes area.
For readability, it reports the behavior of 
 one representative mobile node only. 
The same  trend is observed for all mobile nodes, 
 and all the data required to generate the corresponding charts for each of them 
 are available in the public repository.
The results show that the model achieves high accuracy only on the data distribution 
 currently observed by the node and used for training in that time interval. 
Conversely, when the node moves to a different area, the accuracy drops sharply compared to the performance in the previously visited region. This results from the node not having been trained on the data of this new area.
In contrast, in \Cref{fig:moving-node-c2fl}, 
 where the same scenario is run with \approach{}, 
 the accuracy on previously visited areas is substantially preserved across region transitions, 
 confirming that the proposed approach effectively mitigates forgetting.

This behavior confirms that, 
 in the considered setting,
 spatial heterogeneity becomes a temporal drift for mobile nodes: 
 as the device sequentially adapts to different non-IID regional distributions, 
 it overwrites previously acquired knowledge. 
Therefore, 
 catastrophic forgetting is not only present in decentralized clustered FL under mobility, 
 but also has a substantial impact on the ability of mobile devices to retain useful
 knowledge across areas.

\begin{figure}
    \centering
    \includegraphics[width=0.7\columnwidth]{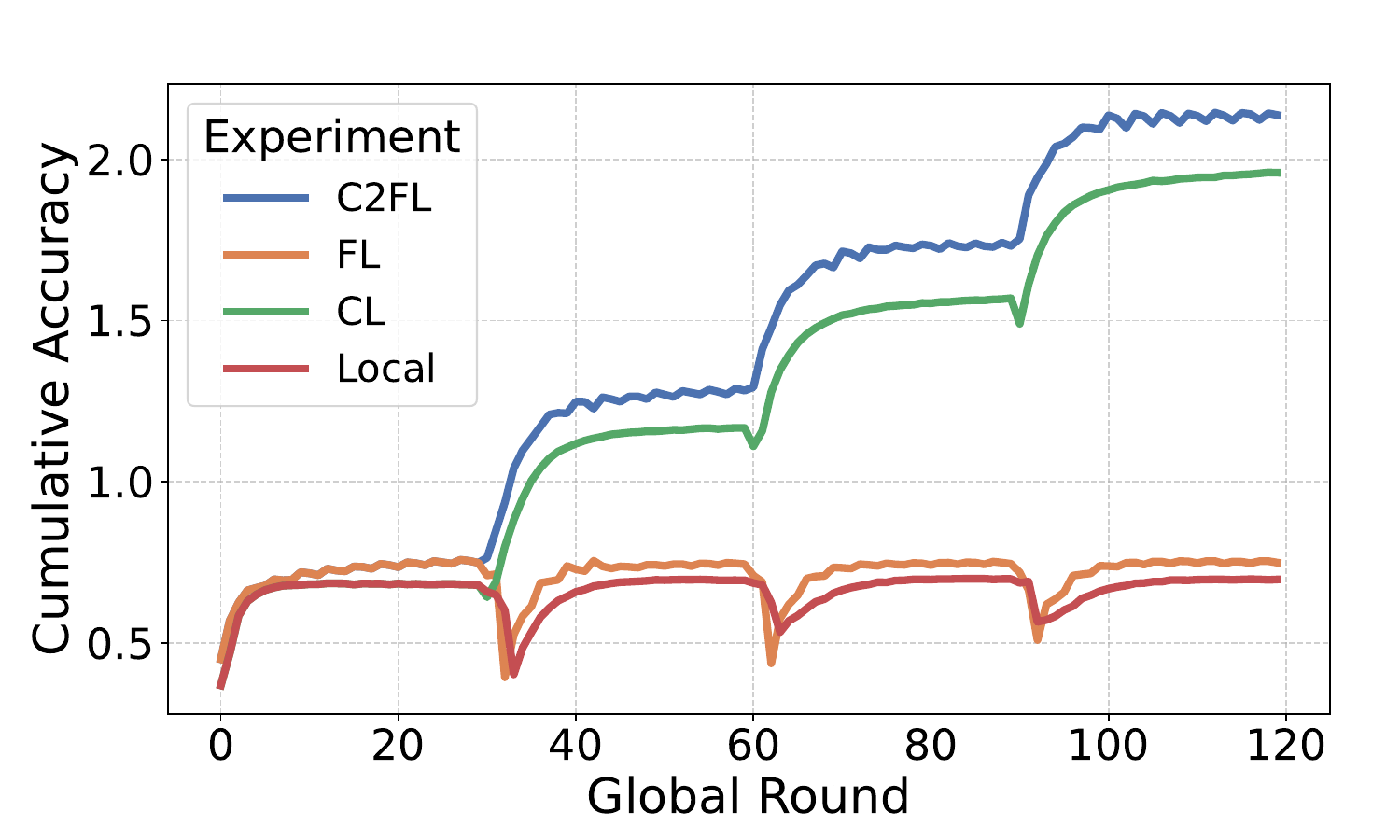}
    \caption{
    Cumulative accuracy $\mathrm{CAcc}$ of the evaluated methods over global rounds.
    The proposed \approach{} approach improves performance with respect to all
    considered baselines.
    }
    \label{fig:comparison}
\end{figure}

The second experiment evaluates whether the proposed approach can reduce 
 catastrophic forgetting under mobility-induced drift (\ref{itm:rq2}) while also benefiting from the collaborative learning process within each region (local/global knowledge integration). 
To this end, 
 we compare \approach{} against three baselines, 
  which also serve as an ablation study of its main components, namely: 
\begin{enumerate*}[label=(\roman*)]
 \item \emph{Local}, 
  where each mobile device trains only on the data currently sensed in its area; 
 \item \emph{CL}, 
  where the device uses continual learning with replay over the data observed 
  along its trajectory but without federated model integration; and 
 \item \emph{FL}, 
  where the device participates in federated learning using only the data currently
  available in the visited area.
\end{enumerate*} 

\Cref{fig:comparison} reports the cumulative accuracy 
 over the regional test distributions. 
The \emph{Local} and \emph{FL} baselines show limited retention across area transitions, 
 since neither of them includes an explicit mechanism to preserve knowledge acquired 
 in previously visited regions. 
Consequently, 
 their performance mainly reflects the ability to fit the data distribution currently perceived 
 by the mobile device. 
The advantage of \emph{FL} over purely local learning is visible,
 but moderate in this setting, 
 likely because the considered classification task is relatively simple; 
 in more complex sensing tasks, 
 the contribution of collaborative learning is expected to become more pronounced.
The \emph{CL} baseline substantially improves over both \emph{Local} and \emph{FL}, 
 confirming that replay is effective in preserving knowledge from previously visited areas 
 and mitigating the drops caused by mobility.
While, however, it does not benefit from the regional consensus, which can be observed in the first 30 rounds, 
 where the performance is lower than \emph{FL} due to the lack of collaborative learning.
Finally, 
 \approach{} achieves the best performance among all evaluated methods.
This shows that combining replay-based continual learning with federated model integration 
 allows mobile nodes to retain past knowledge while also benefiting from the experience 
 of the devices currently located in the same area. 
The adaptive integration of the regional consensus therefore improves 
 and accelerates the adaptation of mobile devices, 
 providing a positive answer to \ref{itm:rq2}.

\section{Conclusions and Future Work}\label{sec:conclusion}

This paper addressed decentralized learning 
 in mobile collective adaptive systems characterized by 
 spatially structured non-IID data 
 and mobility-induced temporal drift. 
We showed that, 
 when devices move across different regions, 
 standard decentralized clustered FL suffers from catastrophic forgetting, 
 as mobile nodes progressively overwrite knowledge acquired in previously visited areas (RQ1). 
To mitigate this issue, 
 we propose \approach{}, 
 a decentralized clustered continual federated learning approach 
 that combines self-organizing spatial clustering, federated model integration, 
 replay-based local training, 
 and adaptive averaging. 
The experimental results show that \approach{} 
 improves knowledge retention and achieves higher cumulative accuracy 
 than all considered baselines balancing local and global objectives (RQ2).

Future work will extend the evaluation along three main directions.
First,
 we plan to validate the approach on additional and more complex datasets,
 including scenarios closer to real-world sensing tasks. 
Second, 
 we aim to integrate and compare multiple continual learning strategies beyond replay,
 such as regularization-based and hybrid methods, 
 to better understand their trade-offs in decentralized clustered FL. 
Third, 
 we will study smoother mobility-induced distribution shifts. 
In the current setting, 
 moving from one area to another causes an abrupt change in the observed distribution;
 future experiments will consider gradual transitions, 
 where the data distribution evolves progressively across space, 
 providing a more realistic model of environmental change.

\section*{Acknowledgments}
The authors have removed acknowledgments to preserve double-blind review requirements. 
Acknowledgments will be added again upon acceptance of the paper.
The authors used Gemini Pro 3.1 (Google) 
 and ChatGPT 5.5 Edu (OpenAI) for grammar 
 and language editing assistance throughout the manuscript. 
All AI-assisted text was reviewed and edited by the authors, 
 who take full responsibility for the content of this paper.

\bibliographystyle{IEEEtran}
\bibliography{biblio}
\end{document}